\documentclass{article}
\usepackage{spconf,amsmath,epsfig}
\usepackage{hyperref}
\usepackage[backend=bibtex, sorting=none]{biblatex}
\usepackage{graphicx}
\usepackage{amsmath}
\usepackage{amssymb}
\usepackage{multirow}
\usepackage{multicol}
\usepackage{bm}
\usepackage{xcolor}
\usepackage{adjustbox}
\usepackage{booktabs}
\usepackage[T1]{fontenc}
\usepackage[utf8]{inputenc}
\title{HyRet-Change: A Hybrid Retentive Network for Remote Sensing Change Detection}
\name{Mustansar~Fiaz$^1$, Mubashir~Noman$^2$, Hiyam Debary$^1$, Kamran Ali$^3$, Hisham Cholakkal$^2$ 
}
\address{$^1$IBM Research, UAE \hspace{1.5mm} $^2$MBZUAI, UAE \hspace{1.5mm} $^3$FAST National University, Karachi, Pakistan} 

\begin{document}
\ninept
\topmargin=0mm
\maketitle
\begin{abstract}

Recently convolution and transformer-based change detection (CD) methods provide promising performance. However, it remains unclear how the local and global dependencies interact to effectively alleviate the pseudo changes. Moreover, directly utilizing standard self-attention presents intrinsic limitations including governing global feature representations limit to capture subtle changes, quadratic complexity, and restricted training parallelism. To address these limitations, we propose a Siamese-based framework, called  HyRet-Change, which can seamlessly integrate the merits of convolution and retention mechanisms at multi-scale features to preserve critical information and enhance adaptability in complex scenes. Specifically, we introduce a novel feature difference module to exploit both convolutions and multi-head retention mechanisms in a parallel manner to capture complementary information.  Furthermore, we propose an adaptive local-global interactive context awareness mechanism that enables mutual learning and enhances discrimination capability through information exchange. We perform experiments on three challenging CD datasets and achieve state-of-the-art performance compared to existing methods. Our source code is publicly available at \url{https://github.com/mustansarfiaz/HyRect-Change}.

\end{abstract}
\section{Introduction}
\label{sec:intro}

Remote sensing change detection (RSCD) is a primitive task where the objective is to categorize each pixel as either changed or unchanged in a given pair of satellite images captured at different times for the same geographical area \cite{shi2020change,  aleissaee2023transformers}. It has various applications including environmental monitoring \cite{fonseca2021pattern}, urban planning \cite{yin2021integrating}, land cover mapping as well as disaster assessment \cite{kucharczyk2021remote}. It is a challenging task due to various complexities characterized by noise, e.g. shadows and variations due to seasons and environmental factors \cite{noman2024remote}.

Traditional approaches involve feature difference techniques in conjunction with a threshold, leading to a binary mask \cite{rosin1998thresholding, rosin2003evaluation}. Nonetheless, such a methodology often encounters challenges like seasonal and lighting variations among distinct image regions, thus impeding the ability to discriminate between semantic and noisy/pseudo changes. Consequently, alternative methods employing hand-crafted features and classifiers such as decision trees \cite{im2005change}, change vector analysis \cite{bovolo2011framework}, support vector machine \cite{volpi2013supervised}, and clustering approaches \cite{aiazzi2013nonparametric, shang2014change}, have been proposed to improve discrimination. Nevertheless, these traditional approaches are subject to various limitations (such as brightness, illumination variations, pose/view variations, and appearance variations) and may struggle to capture comprehensive feature representations.

Recent advances in deep learning techniques including convolutional neural networks (CNNs) and transformers-based CD methods \cite{daudt2018fully, chen2020dasnet, liu2020building,
 li2022transunetcd, , chen2021remote_bit, noman2024remote, fang2021snunet,  yan2022fully_ftn, bandara2022transformer} demonstrate promising performances compared to the conventional CD approaches. 
The convolutional local features emphasize the fine-grained regions \cite{daudt2018fully, chen2020dasnet, liu2020building}, whereas transformer-based global features excavate the relationship between the semantic change-region and background objects \cite{bandara2022transformer, noman2024remote, fang2021snunet}. 
These approaches primarily rely on Siamese architecture, where a shared network generates bi-temporal features and passes them to a difference module to compute the distance between the corresponding features and measure the change intensity maps.
For instance, BIT \cite{chen2021remote_bit} and TransUNetCD \cite{li2022transunetcd} 
employ ResNet \cite{kaiming2016_resnet} backbone to extract features, fuse these feature representations utilizing self-attention to encode the global representation, and input to a difference module to capture the semantic changes between the two-input images.
On the other hand, ChangeFormer \cite{bandara2022transformer} utilizes a Siamese-based self-attention encoder to enlarge the receptive field and extract features at different scales. Later, these encoded features are fused and fed to a convolutional decoder to capture the change regions between the two images. 

Nevertheless, existing methods utilize convolutional or self-attention operations 
exhibit limitations to directly utilizing them in the CD framework. For example, self-attention may reduce the CD performance due to the dominant global contextual information, lack of spatial or local contextual information, and limited capability to generalize well on small CD datasets.
Moreover, self-attention operation while modeling global information exhibits quadratic computational complexity, requires immense  training data and time due to a lack of inductive biases, and has restricted training parallelism.
Whereas,  convolutional operations are effective in capturing the textural and fine details but may face hinderness in encoding the larger changes due to a limited receptive field. Furthermore,  an interaction between global and local to capture both coarse and fine detailed information may further improve the CD performance.
Hence, it is desirable to design such a local-global awareness mechanism that can benefit from both convolution and self-attention in a fully interactive manner, meanwhile uncovering the intricate relationship to raise the model's ability  to capture both subtle and large semantic changes in the scene.

In this paper, we propose an end-to-end Siamese-based CD framework, dubbed HyRet-Change, that aims to capture the local-global context information as well as build an adaptive interaction between them in context-aware ways to precisely identify semantic changes, at multi-scale resolutions, 
in a pair of satellite images taken at distinct time stamps.
The main focus of our paper is to introduce a hybrid feature difference module (FDM) that aims to fuse the merits of CNNs and self-attention in a parallel fashion and present a more cohesive interaction mechanism to prevent crucial information loss and increase the adaptability in complex scenes.
In response to the limitations of self-attention, retentive networks are explored  \cite{sun2307retentive, fan2023rmt} which aim to process the sequences of tokens in parallel representation, recurrent representation, and chunkwise recurrent representation. The main difference between self-attention and retention lies in integrating decay masks. These masks are utilized to regulate the attention weights of individual tokens for the subsequent tokens. Moreover, retention networks process a sequence of tokens instead of individual ones which reduces the computational complexity. In addition to global representation, we intend to exploit local spatial context using depth-wise convolution.  
Furthermore, inspired by FAT \cite{fan2023lightweight}, we integrate an adaptive local-global interaction that can efficaciously enable mutual learning through information exchange in the bi-temporal RS input images. This interaction of long-range dependencies along with the fine-grained key features, thereby hamper the influence of the noisy/pseudo changes while modeling the semantic change regions in CD scenarios. 
 In summary, our contributions are:
\begin{itemize}
\item We propose a hybrid  plug-and-play feature difference module (FDM) to explore rich feature information 
utilizing both self-attention and convolution  operations in a parallel way. This unique integration, at multi-scale features, leverages the
advantages of both local features and long-range contextual information.  
\item We introduce a retention mechanism in our novel FDM to mitigate the limitations of standard self-attention.
\item We introduce an adaptive interaction between local and global representations to exploit the intricate relationship contextually to strengthen the model’s ability to perceive meaningful changes while reducing the effect of pseudo-changes.
\item Our extensive experimental study over three challenging CD datasets demonstrates the merits of our approach while achieving state-of-the-art performance. 
\end{itemize}

\begin{figure}[t!]
\centering
 \includegraphics[width=0.5\textwidth]{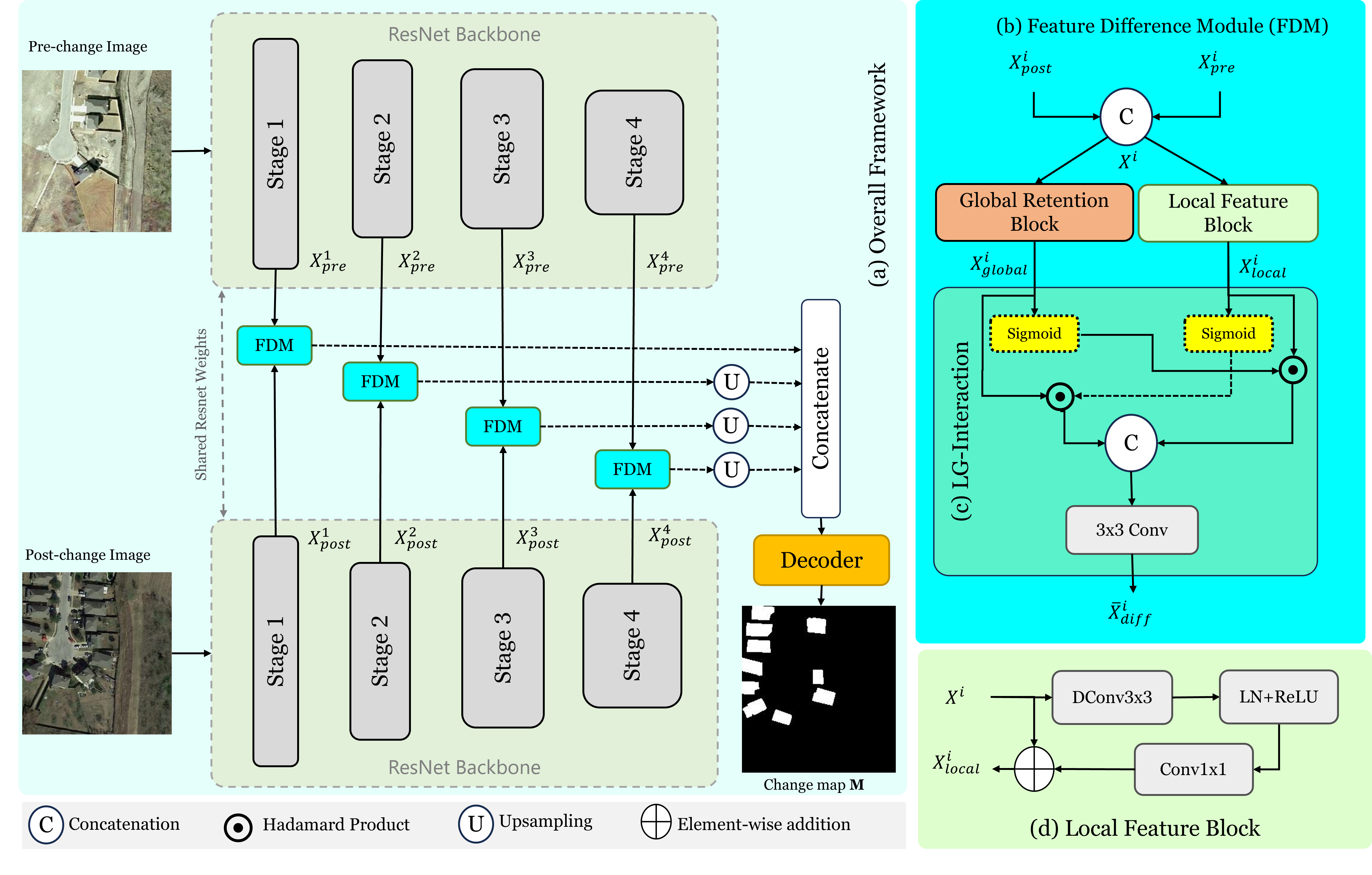} 
\caption{ The (a) 
 illustrates the overall architecture of our proposed CD framework, referred as \textit{ HyRet-Change}. It requires pre-change ($I_{pre}$) and post-change ($I_{post}$) images and is composed of a Siamese-based backbone network, feature difference modules (FDM), and a decoder. The pre-change ($I_{pre}$) and post-change ($I_{post}$) images are input to ResNet \cite{kaiming2016_resnet} to extract multi-scale features at four different resolutions. The $X^i_{pre}$ and $X^i_{post}$ features for the $i$-th scale resolution are input to a difference module, called FDM, to generate enriched difference feature representations $\bar X^i_{diff}$ for the change regions. FDM focuses on exploiting local and global feature representations using the proposed local feature block (LFB) and  global retention block (GRB), respectively. The LFB utilizes depthwise convolution to capture local feature representations as depicted in (d).  We also intend to introduce the retention mechanism to capture global representations using the multi-head \textit{retention} operation while mitigating the limitations of the standard self-attention. Furthermore, we exploit inherent relationships between local and global feature representations to better estimate the subtle and large change regions, which results in enhanced difference features $\bar X^i_{diff}$. These enriched features at different scales are upsampled and fused via concatenation operation and input to a decoder to finally predict the change map $\bm{M}$.
 }
 \label{fig:main_framework}
\end{figure}

\section{Method}
\label{sec:method}
\subsection{Baseline Framework}
\label{ssec:baseline}
Our baseline framework is based on a recent BIT \cite{chen2021remote_bit} method, which provides promising performance, and comprises a Siamese-based feature extractor, a transformer-based encoder block, and a difference module. It utilizes
ResNet \cite{kaiming2016_resnet} backbone to extract features for the input images. These bi-temporal features are fused and input to a transformer block to enhance their receptive field and capture global feature representations. Finally, these rich features are split and an absolute difference operation via convolution operation is performed to estimate the change regions in the RS images.

\noindent \textbf{Limitations: } 
As mentioned earlier, the base framework employs self-attention, focusing on the global change regions compared to local change regions. Additionally, standard self-attention operation exhibits quadratic complexity, lacks parallel token processing, requires larger volumes of training data to generalize well due to the absence of inducive biases, and struggles to capture subtle change regions due to dominant global representations which limits its practicality.  
To tackle the aforementioned challenges, we propose a hybrid feature difference module that strives to identify meaningful differences using both local and global feature representations. 
We intend to exploit the retention concept for the CD task to handle the intrinsic constraints of self-attention and yet achieve better performance.
Furthermore, we focus on exploiting the synchronous interaction between local and global change regions to enhance the discriminative ability to capture both the subtle and large change regions.  
In addition, inspired by FPN \cite{lin2017feature},
we intend to exploit rich feature information at multi-scale resolution to effectively handle the scale variations thereby improving the CD performance.

\subsection{Overall Architecture}
\label{ssec:overall_arch}
The Fig. \ref{fig:main_framework}-(a) illustrates the overall framework of our proposed method, debudded as  HyRet-Change. It comprises of Siamese-based backbone, feature difference modules, and a decoder. The proposed framework requires pre-change $I_{pre}$ and post-change $I_{post}$ images, and input to a Siamese-based feature extractor. We utilize ResNet \cite{kaiming2016_resnet} as a backbone network to extract multi-scale features at four different resolutions (such as $X_{pre}^i$ and $X_{post}^i$ where $i \in {1,2,3,4}$). 
These extracted features at each scale are input to our novel feature difference module (FDM) to generate the rich feature representations $\bar X^i_{diff}$. Later, these enriched features are fused after upsampling at the same resolution as of first scale feature resolution and input to a decoder. The decoder consists of two transpose convolution layers followed by a residual convolutional block. It is responsible for upsampling the features to get the same spatial resolution as the input images and finally estimates the change map $\bm M$.

\subsection{Feature Difference Module (FDM)}
\label{ssec:fdm}
Here, we discuss our novel difference module, comprised of  local feature block (LFB), global retention block (GRB), and  local-global (LG)-interaction module,  that strives to capture both local and global context information and their intrinsic relationship. In contrast to the base framework, we employ concatenation operation instead of difference operation to learn the better change feature representation.
As shown in Fig. \ref{fig:main_framework}-(b),  the input $X_{pre}^i$ and $X_{post}^i$ features at $i$-th scale are concatenated and passed to LFB and  GRB to capture both local and global dependencies, respectively. The attended outputs capturing both local and global representations are input to the LG-interaction module to exploit their intrinsic relationship to better capture the difference feature representations $\bar X^i_{diff}$.

\noindent\textbf{ Global Retention Block (GRB):}\label{ssc_grb}
As mentioned earlier, although standard self-attention is capable of capturing global representations, it demonstrates inherent limitations such as quadratic computational cost and lack of parallel token processing. Inspired by the retention mechanism \cite{fan2023rmt}, we propose to exploit a multi-head retention operation that brings the temporal decay during the sequence modeling in a recurrent manner. 
Following \cite{fan2023rmt}, we flatten the input features and recurrently apply bidirectional retention operations over the tokens following:
\begin{equation}
\label{eq:1}
o_n =\sum_{m=1}^{n} \gamma^{|n-m|}(Q_ne^{in\theta})(K_me^{im\theta})^\dagger v_m,    
\end{equation}
where $n$ is the number of tokens and the retention operation is applied  using Eq. \ref{eq:2} as: 
\begin{equation}
\label{eq:2}
Retention(X) = (Softmax(QK^T) \odot D^{2d})V, 
\end{equation}
where Q, K, and V are defined using Eq. \ref{eq:3} as:
\begin{equation}
\begin{split}
\label{eq:3}
Q = (XW_Q)\odot \Theta,   K = (XW_K)\odot \bar \Theta,  V = (XW_V) \\
\Theta_n = e^{in \theta},   {D}^{2d}_{nm}=\gamma^{|x_n-x_m|+|y_n-y_m|}.
\end{split}
\end{equation}
Here, the $\bar \Theta$ denotes the complex conjugate of the $\Theta$ and ${D}^{2d}_{nm} \in \mathbb{R}^D$ represents the decay mask which symbolizes the relative distance between token pairs. 

\noindent\textbf{ Local Feature Block (LFB):}
To capture the fine detail of change regions, such as the edges and shapes of the change objects, we propose to employ LFB as shown in Fig. \ref{fig:main_framework}-(d). It integrates a depthwise convolution along with a point convolutional layer using the rectified linear unit (ReLU) activation function.  To further enrich the features, we employ a residual connection to provide a direct pathway from the input to preserve the important features and obtain local $ X^i_{local}$ feature representations.

\noindent\textbf{ Local-Global (LG)-Interaction:}
In addition to focusing on local and global change features, we also introduce  the LG-Interaction mechanism within our FDM intending to discern mutual learning. The local and global information exchange in a contextual manner may lead to better discrimination in the change regions.
To do so, as displayed in Fig. \ref{fig:main_framework}-(c), the local features are realized with the Sigmoid function and multiplied with global features to exploit local2global features and similarly, global features exposed to the Sigmoid function are multiplied with local features to obtain global2local features. Later, these bi-directional attended features are fused via concatenation and realized with the convolution layer to finally obtain the enhanced rich feature representations $\bar X^i_{diff}$.

\begin{table*}[t!]
\centering
\caption{{State-of-the-art comparison on LEVIR-CD, CDD-CD, and WHU-CD datasets in terms of F1, IoU, and OA metrics. Our method demonstrates superiority compared to existing methods and obtains state-of-the-art performance. The best two results are in \textcolor{red}{red} and \textcolor{blue}{blue}.
}}
\label{tbl:comaprison_on_LEVIR_CDD_WHU_CD}
\setlength{\tabcolsep}{12.0pt}
\scalebox{0.95}{
\begin{tabular}{l|ccc|ccc|ccc} \hline
\multicolumn{1}{l|}{\multirow{2}{*}{Method}}   &\multicolumn{3}{c|}{LEVIR-CD}  & \multicolumn{3}{c}{CDD-CD}   & \multicolumn{3}{c}{WHU-CD}  \\  \cline{2-10} 
\multicolumn{1}{l|}{} &   F1   & OA & IoU  & F1   & OA & IoU  & F1   & OA & IoU  \\  \cline{1-1} \hline \hline 
FC-Siam-Diff   \cite{daudt2018fully} & 86.31 & 98.67 & 75.92 & 70.61 & 94.95 & 54.57 & 58.81 & 95.63 & 41.66   \\
FC-Siam-Conc   \cite{daudt2018fully}   & 83.69 & 98.49 & 71.96   & 75.11  & 94.95 & 60.14 & 66.63 & 97.04 & 49.95 \\
DASNet    \cite{chen2020dasnet}  & 79.91 & 94.32 & 66.54 & 92.70 & {98.20} & 86.39 & 70.50 & 97.29 & 54.41  \\
DTCDSCN  \cite{liu2020building} & 87.67 & 98.77 & 78.05 & 92.09 & 98.16 & 85.34  & 71.95 & 97.42 & 56.19 \\
STANet   \cite{chen2020spatial}  & 87.30 & 98.66 & 77.40 & 84.12 & 96.13 & 72.22 & 82.32 & 98.52 & 69.95  \\
SNUNet \cite{fang2021snunet} & 88.16 & 98.82 & 78.83 & 83.40 & 96.23 & 72.11 & 83.50 & 98.71 & 71.67 \\
BIT  \cite{chen2021remote_bit}   & 89.31  & 98.92 & 80.68  & 88.90 & 97.47 & 80.01  & 83.98 & 98.75 & 72.39 \\ 
ChangeFormer  \cite{bandara2022transformer}  & 90.40 & {99.04} & 82.48  & 89.83  & 97.68 &  81.53 & 84.93 & 98.82 & 73.80 \\
TransUNetCD  \cite{li2022transunetcd}  & \textcolor{blue}{91.11}  & -- & \textcolor{blue}{83.67}  & \textcolor{blue}{97.17}  & -- &  \textcolor{blue}{94.50}  &  \textcolor{red}{93.59} & -- &  \textcolor{blue}{84.42} \\
CSTSUNet \cite{wu2023cstsunet} & 90.68 &  \textcolor{blue}{99.06} & 82.96 & {96.64} & {99.20}  & 93.50 & -- & -- & --  \\
ELGCNet \cite{noman2024elgc} & 90.33 &   {99.03} & 82.36 & {96.63} & \textcolor{blue}{99.21}  & 93.48 & -- & -- & --  \\
\hline
\textbf{Ours}  & \textcolor{red}{91.89} & \textcolor{red}{99.19} & \textcolor{red}{85.00}  &  \textcolor{red}{97.89} &  \textcolor{red}{99.49} &  \textcolor{red}{95.86} &  \textcolor{blue}{93.57} &  \textcolor{red}{99.50} &  \textcolor{red}{87.92}  \\  \hline
\end{tabular}}
\end{table*}

\section{Experimental Section}
\label{sec:exp}

\subsection{Datasets and Evaluation Protocols} In this paper, we perform experiments on three challenging CD datasets to validate the performance of our CD approach. 
Similar to existing CD methods \cite{bandara2022transformer, yan2022fully_ftn}, we evaluate our approach in terms of   \textit{change class} F1-score, \textit{change class} Intersection over Union (IoU) and overall accuracy (OA) metrics. Among these, the change class IoU is the most challenging for the RSCD task.

\noindent \textbf{\textit{LEVIR-CD} \cite{chen2020spatial}:}
is a public dataset consisting of Google Earth 637 high-resolution (0.5m per pixel) images  of size $1024 \times 1024 \times 3$. Following \cite{chen2021remote_bit, bandara2022transformer}, we crop the non-overlapping patches of spatial sices of  $256 \times 256 \times 3$ with default training, validation, and test splits of  7120, 1024, and 2048, respectively.

\noindent  \textbf{\textit{CDD-CD } \cite{Lebedev2018CHANGEDI}:}
is another challenging publically available dataset comprised of different seasonal variations images.  Following \cite{li2022transunetcd}, the non-overlapping patches are cropped of size $256 \times 256 \times 3$. We utilize the default  training, validation, and test splits of  10000, 3000, and 3000, respectively.

\noindent  \textbf{\textit{WHU-CD } \cite{WHUdataset}:} is another building dataset (having different sizes of buildings) that comprises high-resolution (0.075 m) image pairs of size 32507x15354 pixels. Similar to \cite{chen2021remote_bit}, we crop non-overlapping pairs patches sizes of  $256 \times 256 \times 3$ and obtain  training, validation, and test splits of 6096, 762, and 762, respectively.

\subsection{Implementation Details}
We implemented our approach in PyTorch using two A100 GPUs. The proposed approach requires two-input images of resolution  $256 \times 256 \times 3$ and estimates a binary change mask $\bm M$ which is computed through pixel-wise argmax operation along the channel dimension. We utilize pre-trained ResNet50 \cite{kaiming2016_resnet} as our backbone network to extract multi-scale features. We optimized our method using pixel-wise \textit{cross-entropy} loss and \textit{AdamW} optimizer with a weight decay of 0.01 and beta values equal to (0.9, 0.999). We set a batch size of 16 per GPU and trained the model for 200 epochs with a learning rate of 3e-4, which decreases linearly till the last epoch. 

\begin{figure}[t!]
\centering
 \includegraphics[width=0.5\textwidth]{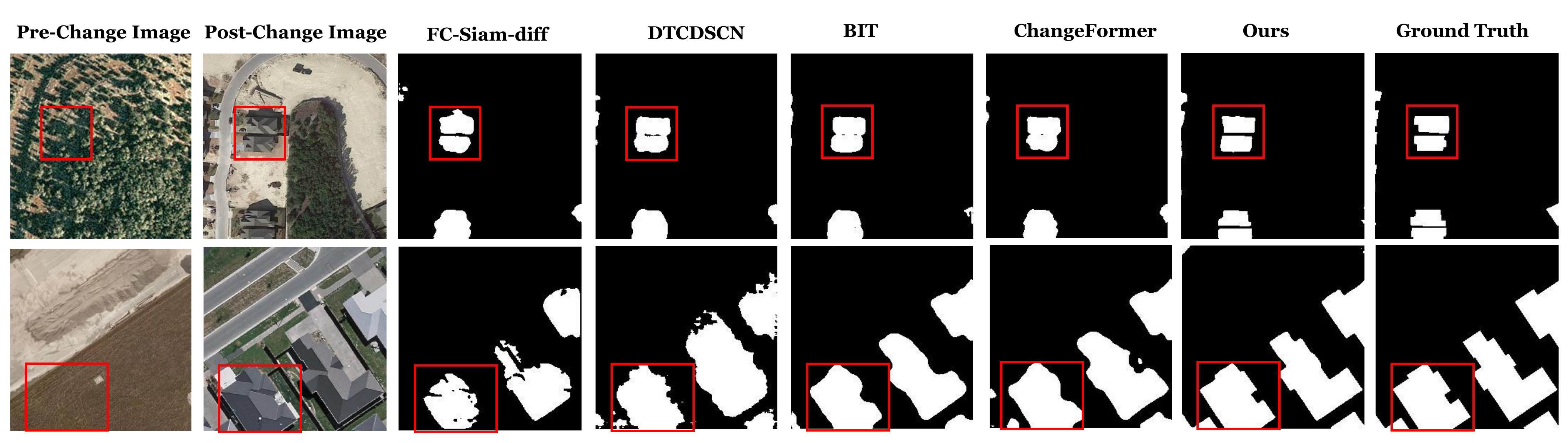} 
\caption{ Qualitative results on challenging examples from the LEVIR-CD (first row) and WHU-CD (second row) datasets. 
The highlighted region in the red box shows that our approach exhibits a better capability to detect both subtle and large change regions with clear boundaries between the pre-and post-change images.}
 \label{fig:qualitative_retChangepng}
\end{figure}

\subsection{State-of-the-art  Comparison}
In Tab.~\ref{tbl:comaprison_on_LEVIR_CDD_WHU_CD}, we present a quantitative comparison with existing state-of-the-art methods over LEVIR-CD, CDD-CD, and WHU-CD datasets in terms of F1, OA, and IoU metrics.
On LEVIR-CD, we notice that among CNN-based approaches, the DTSCDCD \cite{liu2020building} achieves a better IoU of 78.05\%. The  BIT \cite{chen2021remote_bit},  ChangeFormer \cite{bandara2022transformer}, ELGCNet \cite{noman2024elgc}, and TransUNetCD \cite{li2022transunetcd}  are among transformer-based CD methods and obtain IoU of 80.68\%, 82.48\%, 82.36\%, and 83.67\% respectively. We observe that our method achieves IoU of 85.00\% as well as demonstrates state-of-the-art performance in terms of all metrics over the LEVIR-CD dataset.

On the CDD-CD dataset, among both CNN and transformer-based methods, TransUNetCD \cite{li2022transunetcd} achieves an IoU and F1 of 94.50\% and 97.17\%, respectively. Similar to LEVIR-CD, we notice that our method achieves significantly better performance than the best existing methods as shown in Tab.~\ref{tbl:comaprison_on_LEVIR_CDD_WHU_CD} and attains F1, OA, and IoU scores of 97.89\%, 99.49\%, and 95.86\%, respectively. 

We also compare our method over the WHU-CD dataset Tab.~\ref{tbl:comaprison_on_LEVIR_CDD_WHU_CD}. It can be seen that among CNN-based, STANet \cite{chen2020spatial} obtains a 69.95\% IoU score. In contrast,  transformer-based TransUNetCD \cite{li2022transunetcd} presents an F1 and IoU score of 93.59\% and 84.42\%, respectively. 
Our approach presents comparable performance compared to TransUNetCD in terms of F1.
Notably, ours (HyRet-Change) achieves a significant gain of 3.5\% in terms of IoU.  Furthermore, our method achieves better performance in terms of all metrics which demonstrates the proficiency of our method to better capture the local and global representations and their intrinsic relationship.

In Fig. \ref{fig:qualitative_retChangepng}, we also present a qualitative comparison of our method with CNN-based FC-Siam-diff \cite{daudt2018fully}, DTCDSCN \cite{liu2020building} as well as transformer-based BIT \cite{chen2021remote_bit} and ChangeFormer \cite{bandara2022transformer} methods. The qualitative comparison is performed over LEVIR-CD (first row) and WHU-CD (second row) datasets. From Fig. \ref{fig:qualitative_retChangepng}, we observe that ours (HyRet-Change) demonstrates a better capability to discriminative change region features using convolution and retention operations as well as exploit the integral relationship between local and global context representations that complement each other using mutual information exchange learning for the detection of subtle and large change regions.

\begin{table}[t!]
\centering
\caption{Ablation study of our contributions on the LEVIR-CD dataset. 
The best two results are in \textcolor{red}{red} and \textcolor{blue}{blue}, respectively.
}
\label{tbl:ablation_levir}
\scalebox{0.8}{
\begin{tabular}{l|ccc} \hline
Method & F1 & OA & IoU  \\
 \hline \hline
Baseline & 90.71 & 99.01 & 82.85 \\ 
Baseline + MSF  & 91.07 & 99.03 & 83.11 \\
Baseline + MSF + LFB & 91.56 & 99.15 & 84.47 \\  
Baseline + MSF + GRB  & {91.78} & {99.16} & {84.83} \\
Baseline + MSF + LFB + GRB & \textcolor{blue}{91.81}  & \textcolor{blue}{99.17}   & \textcolor{blue}{84.91} \\ \hline
Baseline + MSF + LFB + GRB + LG-Interaction (ours) & \textcolor{red}{91.89}  & \textcolor{red}{99.19}   & \textcolor{red}{85.00} \\ \hline
\multicolumn{4}{l}{
\footnotesize{$^*$MSF represents the multi-scale feature fusion.}}
\end{tabular}}
\end{table}

\subsection{Ablation Study:}
In Table  \ref{tbl:ablation_levir}, we present an ablation study of our method using F1, OA, and IoU measures. We observe that our baseline (row 1) attains 82.85\% IoU score, which utilizes standard self-attention to capture the global feature representations in the difference block. We integrate multi-scale features (row 2) from the backbone network at various scale resolutions to capture better scale variations among the objects. We replace the standard self-attention with our LFB to capture the local features and achieve comparable performance to the baseline. Similarly, we also introduce our global retention block (GRF)  into the baseline framework to mitigate the limitations of self-attention. We notice that retention achieves significant improvement compared to the baseline and achieves an IoU of {84.83}. Later, we integrate both LFB and GRF in row 5 and obtain better improvement in IoU which shows that both local and global feature representations enable better discrimination of the change regions. Finally, row 6 which is our final model obtains the better IoU score of 85.0\%  indicating that mutual learning between local and global feature representations further enhances the model's capacity to capture the better change regions, which shows the merits of our contributions.

\section{Conclusion}
In this work, we propose a Siamese-based RSCD framework that introduces a retention mechanism to enhance the feature discrimination for the change regions between the two input images captured at different time stamps. To be specific, we propose a novel feature difference module that utilizes both local and global context information and strives to capture both subtle and large semantic change regions. We utilize residual depthwise convolution to capture the local dependencies and a multi-head retention operation to encode global dependencies while handling the inherited limitations of standard self-attention. Moreover, we introduce a local-global interaction mechanism to better capture the subtle and large semantic change regions via mutual information exchange. Our experiments on three datasets validate the merits of contributions while achieving state-of-the-art performance compared to the existing CD methods.

\section{Acknowledgments and Collaborations} \label{sec:page}
We thank Levente Klein and Shantanu Godbole for their valuable feedback.


\printbibliography 
\end{document}